\documentclass[twoside,11pt]{article}

\usepackage[preprint]{jmlr2e}

\usepackage{amsmath}
\usepackage{amsfonts}
\usepackage{booktabs}
\usepackage{listings}
\usepackage{textcomp}
\usepackage{xcolor}
\usepackage{tikz}
\usepackage{xurl}
\usepackage{microtype}
\usepackage{float}
\usepackage{caption}
\usepackage{placeins} 
\usetikzlibrary{arrows.meta, shadows.blur, positioning, shapes.geometric, calc} 

\newif\ifArxivBuild
\ArxivBuildtrue 

\jmlrheading{XX}{2025}{X-XX}{XX/XX}{XX/XX}{XXXX-XXXX}{Vincent Koc}

\ShortHeadings{Tiny QA Benchmark\texorpdfstring{$^{++}$}{++}: Micro Gold for LLMOps}{Koc}
\firstpageno{1}

\begin{document}

\title{Tiny QA Benchmark\texorpdfstring{$^{++}$}{++}: Ultra-Lightweight, Synthetic Multilingual Dataset Generation \& Smoke-Tests for Continuous LLM Evaluation}

\author{\name Vincent Koc \email vincentk@comet.com \\
       \addr Comet ML, Inc. \\
       New York, NY, USA
       }

\editor{Editor Name Placeholder} 

\maketitle

\begin{abstract}
\noindent Tiny QA Benchmark\texorpdfstring{$^{++}$}{++} (TQB\texorpdfstring{$^{++}$}{++}) is an ultra-lightweight evaluation suite designed to expose critical failures in Large Language Model (LLM) systems within seconds, contrasting with extensive benchmarks like MMLU or BIG-Bench. At its heart lies a $<$20KB golden dataset of 52 hand-crafted English Question-Answering (QA) triples, ideal for rapid CI/CD checks and prompt engineering. This paper details the evolution into Tiny QA Benchmark\texorpdfstring{$^{++}$}{++}, which significantly expands upon the original TQB. Key enhancements include: (i) a synthetic, on-demand generation toolkit—a Python LiteLLM script ($<$300 lines)—that produces schema-validated micro-benchmarks in any language, domain, or difficulty, with SHA-256 hashing for provenance; and (ii) pre-built multilingual packs $<$20KB for Arabic (AR), German (DE), English (EN), Spanish (ES), French (FR), Japanese (JA), Russian (RU), Korean (KO), Portuguese (PT), Turkish (TR), and Chinese (ZH), enabling immediate cross-lingual smoke tests. We position TQB\texorpdfstring{$^{++}$}{++} as the LLM analogue of software unit tests. Empirically, top-tier models achieve high accuracy ($\approx$ 90\% Exact Match) on the core English set, while performance significantly varies for low-resource languages, demonstrating TQB\texorpdfstring{$^{++}$}{++}'s utility in rapidly detecting regressions or quality shifts in LLMOps workflows. The dataset, generator script, and related tools are released under open-source licenses (see Section~\ref{sec:ethics_license} for details) and hosted on the Hugging Face Hub (\url{https://huggingface.co/datasets/vincentkoc/tiny_qa_benchmark_pp}) and GitHub (\url{https://github.com/vincentkoc/tiny_qa_benchmark_pp}) \citep{koctinyqabenchmarkpp}, promoting accessible and continuous quality assurance in modern LLMOps.
\end{abstract}

\begin{keywords}
  LLM Evaluation, QA Benchmarks, Synthetic Data, LLMOps, Continuous Integration, Smoke Testing, Croissant Metadata
\end{keywords}

\vspace{0.5\baselineskip}
\section{Introduction and Motivation}\label{sec:intro}
Large Language Models (LLMs) are typically evaluated on extensive benchmarks like \emph{MMLU} (Massive Multitask Language Understanding) \citep{hendrycks2021measuringmassivemultitasklanguage} and \emph{BIG-Bench} \citep{srivastava2023imitationgamequantifyingextrapolating}, which cover dozens or even hundreds of tasks with thousands of queries. These comprehensive evaluations demand significant time and compute resources; for example, evaluating a single model across all 57 tasks of MMLU or 204 tasks of BIG-Bench can cost many GPU-hours or API fees. In fast-paced development and deployment of LLM-powered applications, often termed \emph{LLMOps} (the DevOps‑inspired practice of deploying, monitoring, and iterating on LLM applications) teams need something far lighter to guard continuous integration (CI) / continuous deployment (CD) pipelines and interactive development loops. 

This need for rapid feedback became particularly apparent during projects involving iterative prompt optimization, such as the development of prompt engineering solutions for Comet's Opik Optimizer \citep{cometopikoptimizerdocs}, where repeatedly running large evaluation suites proved to be a significant bottleneck, highlighting the demand for quick yet realistic feedback mechanisms.

\paragraph{The Need for Tiny Benchmarks.} The original Tiny QA Benchmark (TQB) \citep{koctinyqabenchmark_original} was created to address this gap by providing a minimal English QA set for rapid CI/CD validation and prompt debugging. Analogous to software unit tests, TQB provides a quick, low-cost signal for basic regressions or integration errors (e.g., broken prompt formats, retrieval failures) by covering elementary general knowledge questions that competent models should answer correctly. Its small size (<17KB) allows instant loading and evaluation in seconds, making it a practical first-line check before invoking resource-intensive benchmarks. While effective as an initial sanity check, TQB intentionally does not aim to replace rigorous benchmarks or provide statistically significant model ranking.

\paragraph{Limitations of a Fixed Core Set.}  While the original TQB performs admirably as a baseline health check, its fixed, English-only nature presents limitations in real‑world deployments which are often multilingual, domain‑specific, and may have unique regulatory constraints. A static set may miss regressions in, for example, French legal chatbots or Turkish banking agents. Furthermore, many teams require the ability to maintain private, scenario‑tailored tests that cannot be easily open‑sourced or derived from a fixed public benchmark.

\paragraph{Contributions of TQB\texorpdfstring{$^{++}$}{++}.} To address these limitations and enhance utility for broader LLMOps, we present \textbf{TQB\texorpdfstring{$^{++}$}{++}}. Its key contributions and design goals are:
\begin{enumerate}
  \item \textbf{Keep the gold standard.} The 52‑item English core remains immutable for deterministic regression.
  \item \textbf{Enable synthetic customisation.} Provide a simple, licence‑compliant generator to mint tiny datasets on demand for any language or any topic.
  \item \textbf{Standardise metadata.} Package artefacts in \emph{Croissant} JSON‑LD \citep{akhtar2024croissant} so tools and search engines can discover and auto‑load them.
  \item \textbf{Promote Open Science.} Release all code, the core dataset, and generation tools under open-source licenses (see Section~\ref{sec:ethics_license} for details), fostering community adoption and extension.
  \item \textbf{Align with LLMOps.} Document CI/CD integration, prompt‑engineering workflows, cross‑lingual drift detection, and observability dashboards.
\end{enumerate}
Together these features deliver a practical, extensible resource for continuous evaluation. By providing an easily accessible, standardized mini-dataset hosted on the Hugging Face Hub (\url{https://huggingface.co/datasets/vincentkoc/tiny_qa_benchmark_pp}) and GitHub (\url{https://github.com/vincentkoc/tiny_qa_benchmark_pp}) under open-source licenses (see Section~\ref{sec:ethics_license} for details) \citep{koctinyqabenchmarkpp}, we aim to encourage the community to adopt more robust testing and CI practices for LLM-based applications.

\subsection{Ethics, Licensing, and Provenance}\label{sec:ethics_license}
The TQB core dataset questions and answers were hand-crafted by the dataset creator from well-known public-domain facts, representing canonical pieces of knowledge that are unlikely to be incorrect or disputed. Since the content is drawn from general knowledge, there are no privacy or sensitivity concerns. The project utilizes a multi-licensing strategy for its components: 
\begin{itemize}
    \item The core English dataset (\texttt{core\_en}), all software (including the generator script), and the Python client library (\texttt{tinyqabenchmarkpp} on PyPI) are licensed under Apache-2.0.
    \item Synthetically generated dataset packs, such as those provided on the Hugging Face Hub or generated by users for distribution, are licensed under a custom ``Eval-Only, Non-Commercial, No-Derivatives'' license.
    \item Croissant JSON-LD metadata files are available under CC0-1.0.
\end{itemize}
The TQB\texorpdfstring{$^{++}$}{++} generator stamps each synthetic item with a SHA-256 hash for provenance tracking. Comprehensive licensing information for each component is available in the project's repository. The Python client library is Apache-2.0 (same as core codebase).


\section{Dataset Structure and Schema}\label{sec:schema}
\subsection{Human‑Curated Core (TQB)}
The human‑curated core of TQB consists of 52 question-answer pairs covering general knowledge domains such as geography, history, science (physics, biology, chemistry), mathematics (including basic arithmetic and calculus), technology (computer science), literature, art, logic puzzles, and temporal/calendar trivia. Each example is represented as a JSON object with the following fields:
\begin{itemize}
    \item \texttt{text} (string): The question prompt, e.g., ``What is the capital of France?''
    \item \texttt{label} (string): The gold answer, e.g., ``Paris''.
    \item \texttt{metadata.context} (string): A one-sentence factual statement supporting the answer (serving as a brief evidence or explanation). For example, ``France is a country in Europe. Its capital is Paris.''
    \item \texttt{tags.category} (string): A broad category for the question (e.g., ``geography'', ``history'', ``math'', ``computer science'', etc.). There are roughly 15 distinct categories in the set, spanning STEM fields and the humanities.
    \item \texttt{tags.difficulty} (string): A rough difficulty level of the question (``easy'' or ``medium'' in this dataset; none are labeled ``hard'' as the dataset was designed to run quickly).
\end{itemize}
All questions have concise answers (mostly single words, numbers, or short phrases). The context field typically restates a known fact and includes the answer, making it useful for extractive QA evaluation or as a verification for closed-book answers. For instance, a question ``Who wrote Romeo and Juliet?'' has context ``Romeo and Juliet is a famous play written by William Shakespeare.'' and answer ``William Shakespeare''. About two-thirds of the items are labelled \emph{easy}, and one-third \emph{medium}, reflecting an intentional bias toward catch-all simplicity. There are no ambiguous prompts or trick questions; every answer is unique in the given context.

The entire dataset is a single JSON file (approximately 17 KB, 52 entries). It is trivial to load in Python via the Hugging Face \texttt{datasets} library or via plain JSON parsing. Being so small, it loads almost instantly, and evaluations over this set complete in negligible time (a modern LLM API can answer all 52 questions in a few seconds). This design choice was deliberate to facilitate fast iteration. The dataset's simplicity ensures that virtually all well-performing models should achieve high accuracy, making any failure very conspicuous, aligning with its role as a canary for regressions.

\subsection{Multi‑Lingual Extensions (TQB\texorpdfstring{$^{++}$}{++})}
Using the generator in TQB\texorpdfstring{$^{++}$}{++} releases packs in ten languages: English (EN), French (FR), Spanish (ES), Portuguese (PT), German (DE), Chinese (ZH), Japanese (JA), Russian (RU), Korean (KO), Turkish (TR), and Arabic (AR). Each pack typically contains 50 QA items (e.g., 10 categories × 10 questions per category), though this is configurable. Users can regenerate these packs, extend them with more questions, or create packs for new languages and domains with a single CLI call using the provided toolkit. The aim for these multi-lingual extensions is not necessarily to ensure identical human-perceived difficulty across languages, which can be highly complex \citep{Koc202XNonLatinLLMEvalComet}, but rather to evaluate a model's consistency and capability on conceptually similar tasks that are formulated in different linguistic contexts through a standardized generation process. This approach can help identify language-specific performance disparities or weaknesses in multilingual models \citep{koc2025generativeailargelanguage}.

\section{Synthetic Data Generation Toolkit (TQB\texorpdfstring{$^{++}$}{++})}\label{sec:generator}
\subsection{Design}
\begin{figure*}[h]
  \centering
  \ifArxivBuild
    \includegraphics[width=0.75\linewidth]{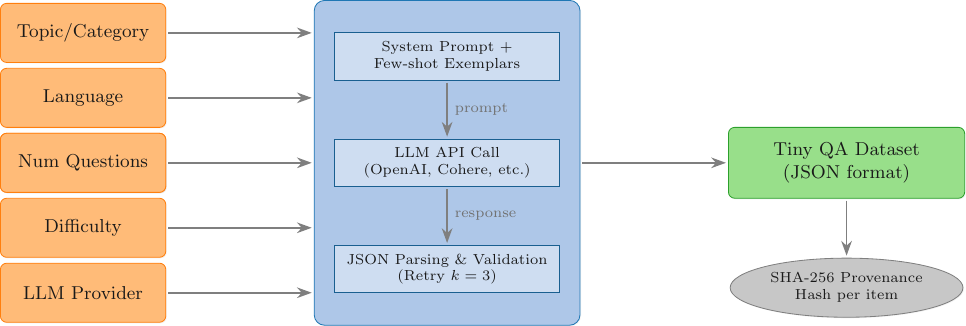} 
  \fi
  \caption{Workflow of the TQB$^{++}$ synthetic data generator, illustrating the process from user inputs (language, topic, difficulty) through LLM-based generation with system prompts and few-shot examples, to JSON validation and final QA item output with provenance tracking.}\label{fig:generator_workflow}
\end{figure*}

A core component of TQB\texorpdfstring{$^{++}$}{++} is its synthetic data generation toolkit. Implemented in approximately 40 lines of Python code using LiteLLM for provider-agnostic LLM calls (see Appendix~\ref{app:generator_code}), the generator can mint bespoke tiny QA datasets. Users can specify parameters such as \texttt{--num} (number of questions), \texttt{--languages} (comma-separated list of language codes), \texttt{--categories} (topics for questions), \texttt{--difficulty}, and \texttt{--provider} (the LLM endpoint to use, e.g., OpenAI, Anthropic, Cohere, or any OpenAI-compatible API). The generation process, illustrated in Figure~\ref{fig:generator_workflow}, involves:
\begin{enumerate}
  \item Crafting a system prompt that instructs the LLM to output structured JSON, adhering to the TQB schema (text, label, context, tags).
  \item Providing two few‑shot exemplars within the prompt to guide the LLM on the desired format and content style.
  \item Sending the generation request to the chosen LLM and parsing the response.
  \item Basic validation of the generated JSON structure, with a retry mechanism (up to $k=3$ attempts) if the output is malformed.
  \item Storing a SHA‑256 hash of each generated item for provenance tracking and reproducibility.
\end{enumerate}

\section{Practical Usage Scenarios}\label{sec:use}
TQB and TQB$^{++}$ are designed to slot easily into various LLMOps and evaluation workflows.

\subsection{CI/CD Pipeline Testing}
The dataset can be used as a unit test for an LLM service. For example, a nightly build or deployment pipeline can automatically run the 52 TQB QA pairs (or a relevant TQB$^{++}$ pack) through the latest model or agent and compare responses to the expected answers. Any incorrect responses can fail the pipeline, alerting engineers before a faulty model is released. A PyTest fixture might load TQB and assert \texttt{exact\_match\_accuracy} $\ge 0.95$. The check runs in approximately 0.5 seconds on a CPU, enabling per‑commit gating. Frameworks like Comet\'s Opik explicitly support this pattern, allowing users to store small eval sets and run them as part of CI (via a PyTest integration), with automated upload of traces and metrics to a monitoring server dashboard.

\begin{figure*}[h]
  \centering
  \ifArxivBuild
    \includegraphics[width=0.55\linewidth]{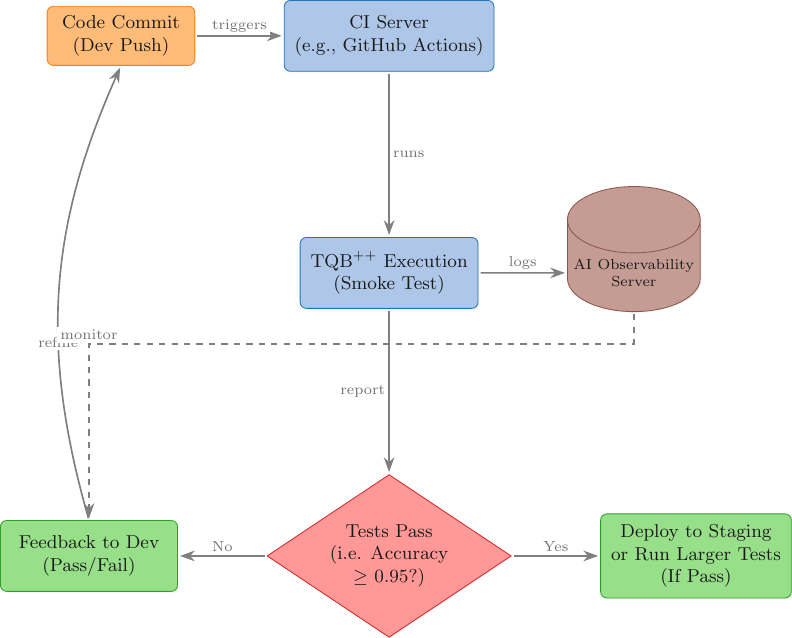}
  \fi
  \caption{Conceptual diagram of TQB\texorpdfstring{$^{++}$}{++} in a CI/CD Pipeline, showing code commit triggering test execution, results logging, decision points, and feedback loops.}\label{fig:cicd_integration}
\end{figure*}

\subsection{Prompt Engineering and Agent Development}
When iterating on prompts or multi‑step agents (e.g., using LangChain or DSPy), developers can rerun the tiny set after each edit. Because the categories are broad, a failure instantly localises issues: e.g., if the math category drops, one might check a calculator tool integration; if history questions fail, a retrieval chain bug might be the cause. If an advanced multi-step agent fails any TQB question that a basic single-call model could handle, that signals a problem in the orchestration. LangChain can directly ingest the Hugging Face dataset object or a converted Pandas DataFrame. DSPy can also utilize these QA pairs as test cases to verify that its compiled strategies yield correct answers.

\subsection{Evaluation Harness Integration}
TQB can be encoded as an OpenAI Evals YAML or a LangSmith dataset, providing dashboards of accuracy over time. Category tags permit fine‑grained tracking (e.g., the science subset of a French TQB$^{++}$ pack). LLM observability platforms increasingly emphasize fine-grained tracing and evaluation \citep{signoz2024llmobservability}, and a fixed tiny test set is a low-noise signal for such monitoring. One could create an eval with TQB to regularly track a model's accuracy on these 52 questions as a baseline health metric \citep{helicone2024prompteval}.

\begin{figure*}[h]
  \centering
  \ifArxivBuild
    \includegraphics[width=0.75\linewidth]{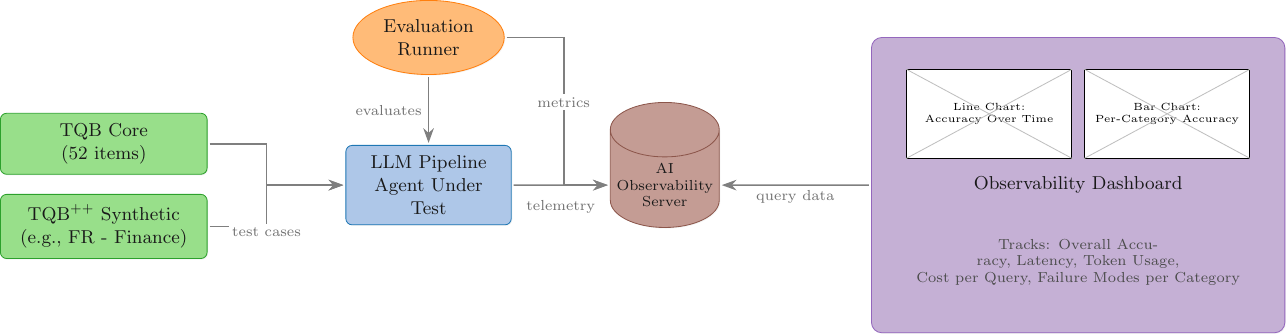}
  \fi
  \caption{Example Observability Dashboard for TQB\texorpdfstring{$^{++}$}{++} Monitoring.}\label{fig:observability_dashboard}
\end{figure*}

\subsection{Cross‑Lingual Drift Detection (TQB\texorpdfstring{$^{++}$}{++})}
By replaying multi‑lingual TQB\texorpdfstring{$^{++}$}{++} packs hourly or daily in production, teams can detect localisation regressions. For example, after an update to a Turkish tokenizer, if accuracy on the TR pack dropped significantly (e.g., 18 percentage points), this would be caught proactively, potentially before widespread user impact.

\subsection{Demonstrations and Adaptive Testing (TQB\texorpdfstring{$^{++}$}{++})}\label{sec:demos_adaptive}
The 52‑item TQB set is small enough for live demos when showcasing new LLMs, toolchains, or evaluation methodologies. It spotlights differences in model capabilities or the effects of prompt changes without overwhelming audiences.

For highly specialized or rapidly evolving domains, TQB\texorpdfstring{$^{++}$}{++} can also support an adaptive testing paradigm through ``test-time dynamic generation''. Instead of relying solely on pre-generated static datasets, new micro-benchmarks can be synthesized on-the-fly, tailored to the specific features, code changes, or even production data drifts being evaluated. For example, when testing a new module that handles a niche topic, a small, highly relevant TQB\texorpdfstring{$^{++}$}{++} pack can be generated just before the test execution. This ensures that the evaluation directly targets the functionality in question with fresh, specific examples. The LiteLLM integration, with its potential for caching similar generation requests, can help manage the latency of such on-demand generation, making it viable for CI/CD environments. This approach moves towards a pseudo ``LLM-as-a-judge'' for dataset creation at the point of testing, offering a higher degree of agility and relevance for complex, fast-moving projects.

\subsection{Monitoring Fine-tuning Dynamics (TQB\texorpdfstring{$^{++}$}{++})}
A key challenge in fine-tuning LLMs is preventing ``catastrophic forgetting'' or unintended erosion of general knowledge and capabilities. TQB\texorpdfstring{$^{++}$}{++}, with its ability to generate targeted micro-benchmarks across diverse categories and languages, offers a lightweight mechanism for monitoring these dynamics. By regularly evaluating a model undergoing fine-tuning against a suite of relevant TQB\texorpdfstring{$^{++}$}{++} packs (both the core English set and synthetic variants for specific domains or languages of interest), developers can gain rapid insights into how the fine-tuning process affects different knowledge areas. For example, a significant drop in accuracy on a ``general science'' TQB\texorpdfstring{$^{++}$}{++} pack after fine-tuning on a narrow legal domain might indicate knowledge erosion. This signal, obtained quickly and cheaply, can inform adjustments to the fine-tuning strategy (e.g., data mixture, regularization) or serve as an early warning before more extensive evaluations are conducted. This approach is particularly valuable for resource-constrained teams or when developing smaller, open-source models where extensive post-fine-tuning evaluations for every epoch or checkpoint are infeasible. The observed log-loss or accuracy degradation on these targeted tiny benchmarks can even be a candidate signal for integration into RLHF processes to help preserve desired general capabilities.

\subsection{Domain-Specific Smoke Tests from Telemetry (TQB\texorpdfstring{$^{++}$}{++})}
Building on adaptive testing, TQB\texorpdfstring{$^{++}$}{++} can be enhanced to generate domain-specific smoke tests by integrating with LLM operational telemetry or domain-specific knowledge bases. For instance, an observability platform monitoring an LLM application might detect emerging themes in user queries or identify specific types of interactions where the LLM struggles. This telemetry could then seed the TQB\texorpdfstring{$^{++}$}{++} generator, prompting it to create a small, highly relevant set of QA pairs focused on these new themes or problematic areas. Such an approach would allow for the automated creation of dynamic, targeted smoke tests that directly validate the system's handling of observed real-world challenges or evolving domain requirements, providing a powerful tool for proactive quality assurance and rapid response to changing operational landscapes.

\subsection{Implications for LLMOps and Future Work}\label{sec:implications_future}
The experimental validation reinforces the potential of TQB\texorpdfstring{$^{++}$}{++} as a practical tool for LLMOps. The ability to quickly generate targeted, multilingual micro-benchmarks that are sensitive to model capabilities, language differences, and question difficulty offers several advantages for continuous evaluation and quality assurance:

\begin{itemize}
    \item \textbf{Efficient CI/CD Gates:} Small TQB\texorpdfstring{$^{++}$}{++} packs can act as rapid, low-cost checks in CI/CD pipelines, catching regressions before deployment without the overhead of large benchmark suites.
    \item \textbf{Agile Prompt Engineering:} Developers can iterate on prompts and agent designs, using relevant TQB\texorpdfstring{$^{++}$}{++} sets for immediate feedback on how changes impact core QA performance across different categories or languages.
    \item \textbf{Targeted Drift Detection:} Custom-generated TQB\texorpdfstring{$^{++}$}{++} packs for specific domains or languages of interest can monitor for performance drift in production systems, as discussed in Section~\ref{sec:use}.
    \item \textbf{Resource-Constrained Evaluation:} Teams with limited access to compute or expensive model APIs can still perform meaningful, albeit coarse-grained, evaluations and comparisons.
\end{itemize}

The possibility of theoretically feeding domain knowledge or LLM call telemetry into the TQB\texorpdfstring{$^{++}$}{++} generation process to create highly contextualized smoke tests, as suggested during our research, presents a pertinent direction for future work. This could involve:
\begin{itemize}
    \item Developing workflows where production data drifts (e.g., new topics appearing in user queries) trigger the generation of fresh, relevant TQB\texorpdfstring{$^{++}$}{++} micro-benchmarks to ensure the system handles these new scenarios correctly.
    \item Integrating the TQB\texorpdfstring{$^{++}$}{++} generator with LLM observability platforms, allowing automated creation of small validation sets based on monitored operational data or identified failure modes. This would enable a form of ``self-healing'' or adaptive evaluation where the test suite evolves with the application and its usage patterns.
\end{itemize}

Further research could also explore more sophisticated synthetic generation techniques, potentially incorporating LLM-as-a-judge mechanisms directly into the generation loop for on-the-fly quality filtering and refinement \citep{Koc2025LeveragingMultipleLLMEvaluators}. Enhancing the generator to allow for more fine-grained control over the distribution of generated question types (e.g., factual, inferential, mathematical) within categories could also increase its utility. The current study focused on `o3-mini` for generation; exploring the quality and characteristics of TQB\texorpdfstring{$^{++}$}{++} datasets generated by other LLMs (including open-source models) would also be a valuable contribution.

Overall, the experiments demonstrate that TQB\texorpdfstring{$^{++}$}{++} provides a flexible and effective solution for lightweight LLM smoke testing, bridging a gap between rapid iteration needs and the requirement for continuous quality assurance in the LLMOps lifecycle.

\section{Evaluation Philosophy}\label{sec:eval}
TQB\texorpdfstring{$^{++}$}{++} follows a \emph{small‑but‑sufficient} ethos for early-stage validation, in contrast to exhaustive benchmarks aimed at fine-grained model differentiation. While the field currently lacks a comprehensive formal theory of ``LLM smoke testing'' or ``minimal competency evaluation,'' TQB\texorpdfstring{$^{++}$}{++} serves as a practical case study and a step towards defining such a framework. Its design and application motivate several desirable properties (desiderata) for effective smoke tests in LLMOps:
\begin{itemize}
  \item \textbf{Rapid Execution Low Cost:} Tests should complete in seconds, incurring minimal computational or financial overhead, to be viable for per-commit checks.
  \item \textbf{High Sensitivity to Basic Flaws:} The benchmark should target foundational capabilities where regressions are clear indicators of systemic problems (e.g., broken prompt formats, context retrieval failures, severe performance degradation).
  \item \textbf{Broad Initial Coverage:} While tiny, the test should span a diverse range of general knowledge or core functionalities to catch a variety of potential issues early.
  \item \textbf{Configurability Extensibility:} The ability to generate targeted variants (e.g., different languages, domains, difficulties, as with TQB\texorpdfstring{$^{++}$}{++}'s generator) allows tailoring smoke tests to specific project needs.
  \item \textbf{Clear Interpretation of Results:} Outcomes should be easily understandable, often binary (pass/fail) or near-binary, facilitating quick decision-making in CI/CD pipelines. TQB's design, where high accuracy is expected, makes deviations highly salient.
  \item \textbf{Deterministic Core with Stochastic Options:} An immutable core set (like TQB's 52 items) ensures consistent regression detection, while synthetic generation (TQB\texorpdfstring{$^{++}$}{++}) allows for randomized variants to combat overfitting to the test set.
\end{itemize}

This approach is about ensuring minimal competence and catching glaring problems. 
\begin{itemize}
  \item \textbf{Binary safety and Signal in Deviations.} The set is designed to be easy enough that any error on the core TQB signals a potentially serious regression. We expect any advanced model to nearly ace this test; thus, even one failure is noteworthy. If a model's performance drops from 98\% to 90\% on TQB, that is a clear signal to investigate. Evaluation is typically done via exact-match or simple string comparison, making automated grading straightforward.
  \item \textbf{Cheap Redundancy with Synthetic Variants.} The TQB\texorpdfstring{$^{++}$}{++} generator allows teams to create unlimited synthetic variants. This enables randomisation of tests, which can help combat overfitting to a fixed small set and tailor evaluations to specific domains or languages not covered by the core set. For multilingual variants, the focus is on assessing a model's consistent performance on tasks generated with similar logic across different languages, rather than strictly calibrated cross-lingual human difficulty, thereby aiding in the detection of language-specific capability gaps (see also \S\ref{sec:schema} and \citep{koc2025generativeailargelanguage, Koc202XNonLatinLLMEvalComet}).
  \item \textbf{Two‑stage Pipeline.} TQB is intended as a first-line check. Models or pipelines should pass tiny tests first, then proceed to more comprehensive and resource-intensive benchmarks (e.g., MMLU, BIG‑Bench, HELM). This mirrors the unit testing → integration testing → performance testing progression in traditional software development.
\end{itemize}
This philosophy is supported by recent findings that tiny targeted benchmarks can be useful. For instance, \citet{polo2024tinybenchmarksevaluatingllmsfewer} show that evaluating an LLM on as few as 100 curated examples can approximate its performance on a much larger benchmark with surprising accuracy for ranking purposes. The goal of TQB is less about estimating full benchmark scores and more about acting as a safety net or a detector of severe issues: coarse but quick. While very small benchmarks can yield high variance and are not statistically reliable for measuring incremental improvements between top-tier models \citep{hochlehnert2025soberlookprogresslanguage}, TQB's purpose is largely binary for basic models: detect disasters, not measure fine-grained triumphs. After passing the TQB test, a model should still be put through rigorous, large-scale evaluations.

\vspace{0.5\baselineskip}
\section{Experimental Setup}\label{sec:experiments}

Performance was primarily assessed using two metrics. For clarity in reporting in this section, all scores (EM and LR) are generally presented on a 0-100 scale (e.g., an EM of 0.904 is reported as 90.4), unless specified otherwise by the original metric definition (e.g. F1-scores which are typically 0-1). When ``LR Score'' or ``LR Accuracy'' is reported in tables (such as Table~\ref{tab:app_detailed_lr_scores}) and general performance discussions, it refers to the percentage of items achieving a rsaw Levenshtein Ratio of 0.75 or higher, effectively representing an accuracy based on this threshold. Raw Levenshtein Ratios themselves are used directly in the threshold calibration analysis (Section~\ref{sec:results}).
\begin{itemize}
    \item \textbf{Exact Match (EM) Accuracy:} As defined in Section~\ref{sec:eval}, after normalizing both predicted and gold answers (lowercase, removal of articles, punctuation, and extra whitespace).
\end{itemize}

\subsection{Metrics and Test Criteria}
For the core TQB (52 items), the primary metric is \textbf{Exact Match (EM) accuracy}. Given a question $q_i$ with gold answer $a_i$ and model prediction $p_i$, EM is defined as:
\[ EM_i = \begin{cases} 1 & \text{if } normalize(p_i) = normalize(a_i) \\ 0 & \text{otherwise} \end{cases} \]
where $normalize(s)$ is a function that typically converts $s$ to lowercase, removes articles (a, an, the), punctuation, and extra whitespace. Overall accuracy is $\frac{1}{N} \sum_{i=1}^{N} EM_i$ for $N=52$.

A pass/fail threshold can be set, e.g., $Accuracy \ge 0.95$ (allowing for $\approx 2$ errors).

While EM is a stringent and unambiguous metric suitable for the deterministic nature of the core TQB, it may penalize responses that are semantically correct but differ slightly in phrasing. For generated TQB\texorpdfstring{$^{++}$}{++} datasets, or for use-cases where minor answer variations are acceptable, alternative metrics can be more appropriate, such as the Levenshtein Ratio (LR). The validation (Section~\ref{sec:overall_perf}) employed both EM and LR. Although LR can offer more nuanced scoring, EM is often preferred for its simplicity and speed in high-frequency smoke testing scenarios, where a clear, binary signal is most valuable. The choice of metric can depend on the specific goals of the evaluation. For instance, \textbf{Levenshtein distance} \citep{levenshtein1966binary} quantifies the minimum number of single-character edits (insertions, deletions, or substitutions) required to change one word into the other. This raw distance is often normalized into a similarity ratio, e.g., $1 - (\text{lev}(a,b) / \max(|a|, |b|))$. For such a normalized ratio, a suitable pass threshold might be lower than for EM, for example, around $0.75$, though an empirically calibrated threshold (see Section~\ref{sec:results}) is recommended. This threshold could be calibrated based on human feedback assessing acceptable answer variations (e.g., initial calibration for TQB based on a sample of $n=5$ human evaluations per item suggested that certain semantic equivalents would pass at this level but fail EM). 

For a deeper semantic understanding, especially with more complex or nuanced answers that might arise in custom-generated TQB\texorpdfstring{$^{++}$}{++} packs, \textbf{embedding-based distance} (e.g., cosine similarity between answer embeddings) can provide a measure of semantic equivalence. If $A$ and $B$ are the $n$-dimensional vector embeddings of the gold answer and the predicted answer respectively, the cosine similarity is defined as:
\[ 	ext{cos}(A, B) = \frac{A \cdot B}{\|A\| \|B\|} = \frac{\sum_{k=1}^{n} A_k B_k}{\sqrt{\sum_{k=1}^{n} A_k^2} \sqrt{\sum_{k=1}^{n} B_k^2}} \]
While these alternative metrics offer valuable flexibility, particularly for the diverse outputs possible with TQB\texorpdfstring{$^{++}$}{++} variants, the core TQB evaluation in this paper primarily relies on EM for its strictness and clarity as a smoke test. These more nuanced metrics are often more practical for multilingual contexts, particularly with non-Latin scripts, compared to n-gram overlap metrics like BLEU or ROUGE which, while common in other NLP tasks, may be less suitable for short, factual answers and can face challenges with morphologically rich languages \citep{Koc202XNonLatinLLMEvalComet, koc2025generativeailargelanguage}.

\vspace{0.5\baselineskip}
\section{Results and Discussion}\label{sec:results}
The models evaluated in these experiments are detailed in Appendix~\ref{app:evaluation_protocol} (Table~\ref{tab:app_models_evaluated}).

\subsection{Overall Performance Trends}\label{sec:overall_perf}
One of the primary goals for TQB\texorpdfstring{$^{++}$}{++} is to serve as a rapid evaluation tool sensitive enough to detect known performance variations, such as those between LLMs of different sizes within the same model family, or across diverse datasets. The experiments aimed to validate this sensitivity. We anticipated that larger, more computationally intensive models would generally outperform their smaller, distilled counterparts, and that performance would vary based on language and the nature of the questions. TQB\texorpdfstring{$^{++}$}{++} should surface these expected gradations quickly.

\begin{figure*}[h] 
  \centering
  \includegraphics[width=0.9\linewidth]{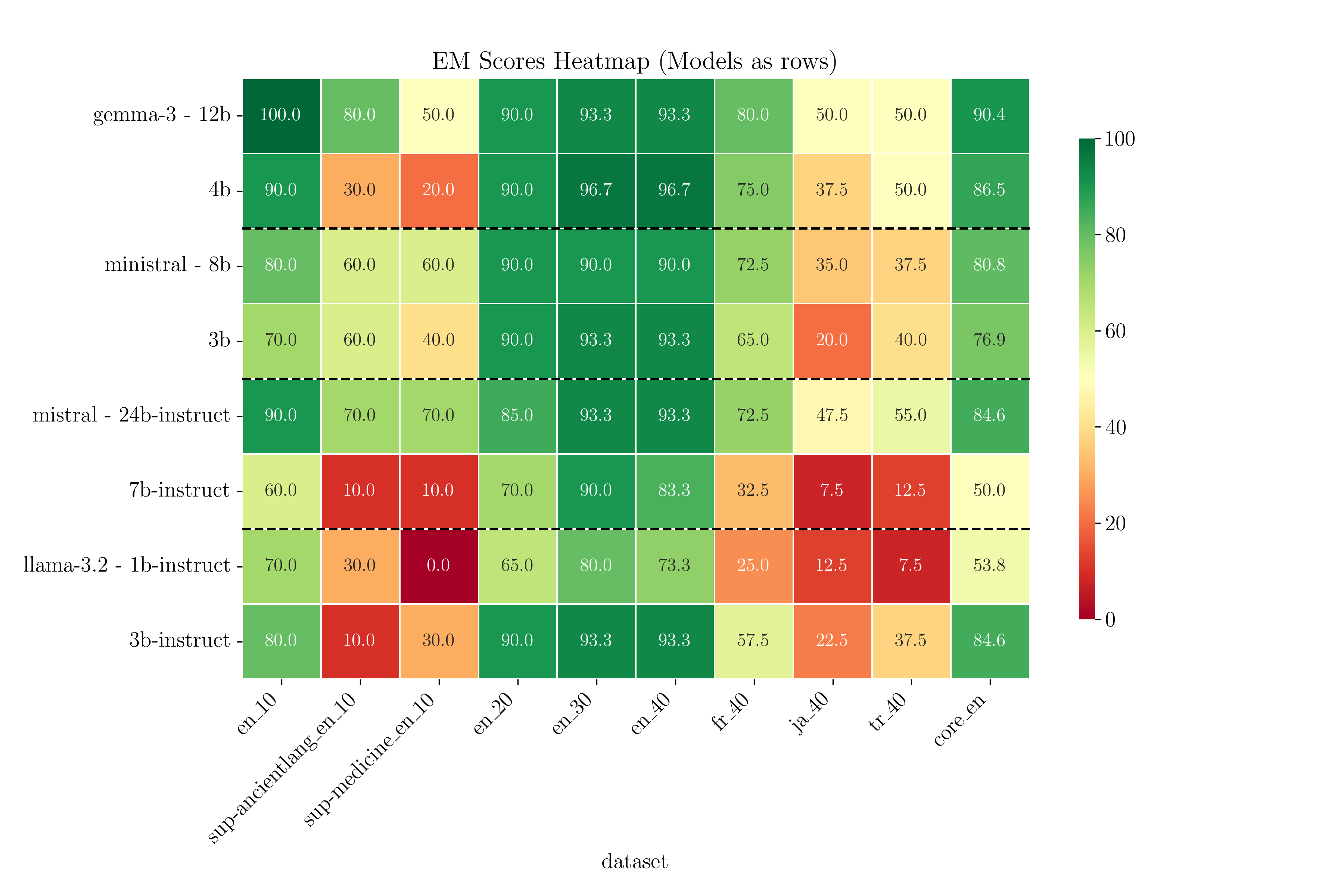} 
  \caption{Heatmap of Exact Match (EM) scores across models and datasets (including supplementary challenge datasets \texttt{sup-ancientlang\_en\_10} and \texttt{sup-medicine\_en\_10}). Darker shades indicate higher accuracy. Models are grouped by family and generally ordered by size/capability within families, illustrating performance degradation with decreasing model weight (e.g., Gemma-3 12b vs. 4b) and variance across datasets.}\label{fig:heatmap_em_scores}
\end{figure*}

The experiments revealed clear performance gradations across models and datasets, underscoring TQB$^{++}$'s ability to differentiate model capabilities even with small test sets. Figure~\ref{fig:heatmap_em_scores} presents a heatmap of EM scores, providing a visual overview. Models are grouped by family (e.g., Gemma, Mistral) and generally ordered by size/capability, allowing for visual inspection of performance hierarchies. A similar heatmap for LR scores (using an empirically determined threshold, see Section~\ref{sec:results}) is provided in Appendix~\ref{app:lr_heatmap_figure} (Figure~\ref{fig:app_heatmap_lr_combo}).
\FloatBarrier

As anticipated, and visibly in Figure~\ref{fig:heatmap_em_scores}, larger and more capable models within the same family generally outperformed their smaller counterparts. For instance, \texttt{gemma-3-12b} consistently scored higher than \texttt{gemma-3-4b} across most datasets (e.g., on \texttt{core\_en}, EM 90.4 vs 86.5; on \texttt{pack\_fr\_40}, EM 80.0 vs 75.0). Similarly, \texttt{mistral-24b-instruct} surpassed \texttt{mistral-7b-instruct} (e.g., on \texttt{core\_en}, EM 84.6 vs 50.0). This trend is a key indicator that TQB\texorpdfstring{$^{++}$}{++} effectively reflects model scale and capability. This pattern is also observed with LR metrics (see Appendix~\ref{app:lr_heatmap_figure}). For a comprehensive breakdown of EM scores for all models across all datasets, including calculated deltas between intra-family model variants, see Table~\ref{tab:app_detailed_em_scores} in Appendix~\ref{app:detailed_scores}. The \texttt{core\_en} dataset (52 human-curated items), generally elicited high scores from the top-performing models, establishing a baseline of expected performance.

The synthetically generated English packs of varying sizes (10, 20, 30, and 40 items, denoted \texttt{pack\_en\_10}, \texttt{pack\_en\_20}, \texttt{pack\_en\_30}, and \texttt{pack\_en\_40} respectively) also showed similar performance patterns, indicating that even small synthetic sets can effectively rank models by capability. Performance on the specialized \texttt{sup-} datasets (ancient languages and medicine, each 10 items) was notably lower across all models, reflecting their \texttt{'hard'} difficulty and niche topics, and demonstrating TQB\texorpdfstring{$^{++}$}{++}'s utility in creating targeted challenging evaluations.

\subsection{Multilingual Performance Analysis}\label{sec:multilingual_perf}
A key goal of TQB$^{++}$ is to facilitate quick assessment of multilingual capabilities. The experiments compared performance on the English datasets (core and synthetic) with newly generated 40-item packs for French (FR), Japanese (JA), and Turkish (TR).

Table~\ref{tab:mean_em_by_lang} summarizes the mean EM scores (averaged across all models, excluding specialized \texttt{sup-} datasets). More revealingly, it also shows EM scores for a specific model pair (\texttt{gemma-3-12b} vs. \texttt{gemma-3-4b}) on representative datasets for each language, highlighting how the performance $\Delta$ between model variants can change across languages. English datasets consistently yielded the highest scores. Japanese and Turkish presented more significant challenges for most models, and the performance $\Delta$ between model sizes can become more pronounced for these languages.

\begin{table}[htbp]
  \centering
  \caption{Mean Exact Match (EM) Scores by Language (0-100 Scale) and Example Model Pair Performance.}\label{tab:mean_em_by_lang}
  \footnotesize 
  \begin{tabular}{@{}lccccc@{}}
    \toprule
    Language & ISO   & Avg. EM      & \texttt{gemma-3-12b} & \texttt{gemma-3-4b} & \textbf{$\Delta$} \\
             & Code  & (all models) & (on 40-52 items)   & (on 40-52 items)   & \textbf{(12b-4b)} \\
    \midrule
    English  & en  & 86.1         & 90.4 (\texttt{core\_en}) & 86.5 (\texttt{core\_en}) & \textbf{3.9} \\
    French   & fr  & 60.0         & 80.0 (\texttt{pack\_fr\_40}) & 75.0 (\texttt{pack\_fr\_40}) & \textbf{5.0} \\
    Japanese & ja  & 29.1         & 50.0 (\texttt{pack\_ja\_40}) & 37.5 (\texttt{pack\_ja\_40}) & \textbf{12.5} \\
    Turkish  & tr  & 36.3         & 50.0 (\texttt{pack\_tr\_40}) & 50.0 (\texttt{pack\_tr\_40}) & \textbf{0.0} \\
    \bottomrule
  \end{tabular}
\end{table}

Further analysis of EM scores by language and category (details available in supplementary materials) reveals nuances. For instance, while some categories might see relatively stable performance across EN and FR, the drop-off can be more pronounced for JA and TR. This degradation is often more severe for smaller model variants within the same family. For example, while \texttt{gemma-3-12b} achieved an EM of 50.0 on \texttt{pack\_ja\_40} and 50.0 on \texttt{pack\_tr\_40}, the smaller \texttt{gemma-3-4b} scored lower at 37.5 and 50.0 respectively. An even smaller model like \texttt{llama-3.2-1b-instruct} scored only 12.5 on \texttt{pack\_ja\_40} and 7.5 on \texttt{pack\_tr\_40}. This suggests that while larger models may possess some zero-shot or few-shot capabilities in these languages, smaller models struggle significantly more, making TQB\texorpdfstring{$^{++}$}{++} effective for detecting such weaknesses rapidly. The performance decay underscores the importance of specific multilingual evaluations, as strong English performance does not guarantee comparable efficacy in other languages, especially those typologically distant from English or less represented in common training corpora.

These findings align with the motivation for TQB\texorpdfstring{$^{++}$}{++}'s multilingual generation capability: providing a simple way to create targeted smoke tests that can flag potential issues in non-English language handling before more exhaustive (and expensive) multilingual benchmarks are run.

\subsection{Impact of Data Characteristics}\label{sec:data_char_impact}
The TQB\texorpdfstring{$^{++}$}{++} framework allows for tagging items with metadata like difficulty and category, enabling more granular performance analysis.

\begin{table*}[htb] 
  \begin{minipage}[b]{0.47\linewidth} 
    \centering
    \caption{EM Scores by Difficulty.}
    \label{tab:em_by_difficulty}
    \footnotesize 
    \begin{tabular}{@{}llccc@{}}
      \toprule
      Model Family & Variant & Easy & Medium & Hard \\
      \midrule
      Gemma-3 & \texttt{gemma-3-12b} & 84.8 & 80.2  & 49.0 \\
              & \texttt{gemma-3-4b}  & 84.8 & 74.5  & 29.4 \\
              & \textbf{\textit{$\Delta$ (12b-4b)}} & \textbf{\textit{0.0}} & \textbf{\textit{5.7}}  & \textbf{\textit{19.6}} \\
      \midrule
      Mistral & \texttt{mistral-24b (it)} & 82.4 & 72.6  & 58.8 \\ 
              & \texttt{mistral-7b (it)}  & 54.4 & 45.3  & 9.8 \\  
              & \textbf{\textit{$\Delta$ (24b-7b)}} & \textbf{\textit{28.0}} & \textbf{\textit{27.3}}  & \textbf{\textit{49.0}} \\
      \bottomrule
    \end{tabular}
  \end{minipage}\hfill
  \begin{minipage}[b]{0.47\linewidth} 
    \centering
    \caption{EM Scores by Category.}\label{tab:em_by_selected_category}
    \footnotesize 
    \begin{tabular}{@{}lccc@{}}
      \toprule
      Variant & Art & Math \\
      \midrule
      \texttt{gemma-3-12b}        & 88.2 & 85.3 \\
      \texttt{gemma-3-4b}         & 82.4 & 79.4 \\
      \textbf{\textit{$\Delta$ (12b-4b)}}    & \textbf{\textit{5.8}} & \textbf{\textit{5.9}} \\
      \midrule
      \texttt{mistral-24b (it)} & 82.4 & 82.4 \\ 
      \texttt{mistral-7b (it)}  & 47.1 & 50.0 \\ 
      \textbf{\textit{$\Delta$ (24b-7b)}}    & \textbf{\textit{35.3}} & \textbf{\textit{32.4}} \\
      \bottomrule
    \end{tabular}
  \end{minipage}
  \caption*{\footnotesize Mean Exact Match (EM) Scores (0-100 Scale). Left (Table~\ref{tab:em_by_difficulty}): by Stated Question Difficulty. Right (Table~\ref{tab:em_by_selected_category}): by Selected Question Categories. (it) denotes instruct-tuned models.}
\end{table*}

\paragraph{Performance by Difficulty.} As expected, model performance generally decreased as the stated difficulty of questions increased from \texttt{'easy'} to \texttt{'medium'} to \texttt{'hard'}. Table~\ref{tab:em_by_difficulty} illustrates this trend for EM scores, highlighting how TQB\texorpdfstring{$^{++}$}{++} can detect performance variations even between models of the same family. Most models maintained higher accuracy on \texttt{'easy'} questions, but scores typically dropped for \texttt{'medium'} and more significantly for \texttt{'hard'} questions. This differentiation is noticeable even between closely related models; for example, the $\Delta$ between \texttt{gemma-3-12b} and its smaller counterpart \texttt{gemma-3-4b} widens as questions become harder, a key sensitivity TQB\texorpdfstring{$^{++}$}{++} aims to expose. This demonstrates that the synthetic generator, guided by simple difficulty prompts, can produce datasets that elicit differential performance reflective of question complexity. The two supplementary datasets (\texttt{sup-ancientlang\_en\_10} and \texttt{sup-medicine\_en\_10}), which were entirely composed of \texttt{'hard'} questions, proved challenging for all models (scores often below 50.0, refer back to Figure~\ref{fig:heatmap_em_scores}).

\paragraph{Performance by Category and Generator Bias.} Analyzing scores by category (full details in supplementary materials, with a summary heatmap in Appendix~\ref{app:lr_heatmap_figure} (Figure~\ref{fig:app_heatmap_category_combo})) reveals that model performance can vary significantly across different knowledge domains. These patterns can indicate relative strengths or weaknesses in a model's training data or reasoning capabilities for specific topics. Table~\ref{tab:em_by_selected_category} shows EM scores and deltas for illustrative categories for the same model pairs, again showing TQB\texorpdfstring{$^{++}$}{++}'s ability to surface fine-grained differences.

\subsection{Levenshtein Ratio Calibration Details}
While Exact Match (EM) is a stringent metric, it can be overly punitive for answers that are semantically correct but differ slightly in phrasing or due to minor OCR-like errors in model outputs. The Levenshtein Ratio (LR) offers a more flexible alternative. However, choosing an appropriate acceptance threshold for LR is crucial. An LR threshold that is too low might accept incorrect answers, while one that is too high approaches EM in strictness.

To determine a data-driven optimal LR threshold, a bootstrapping analysis was performed. For each potential LR threshold from 0.00 to 1.00 (in increments of 0.01), EM scores were treated as ground truth (1 for a match, 0 otherwise). Precision, recall, and F1-score were calculated by comparing LR-based classifications (answer considered correct if its LR score $\ge$ threshold) against these EM-based ground truths across all model predictions on all datasets. This process was repeated with 1000 bootstrap samples to establish confidence intervals, although the primary interest was the threshold maximizing the F1-score.

\begin{figure}[H] 
  \centering
  \includegraphics[width=1\linewidth]{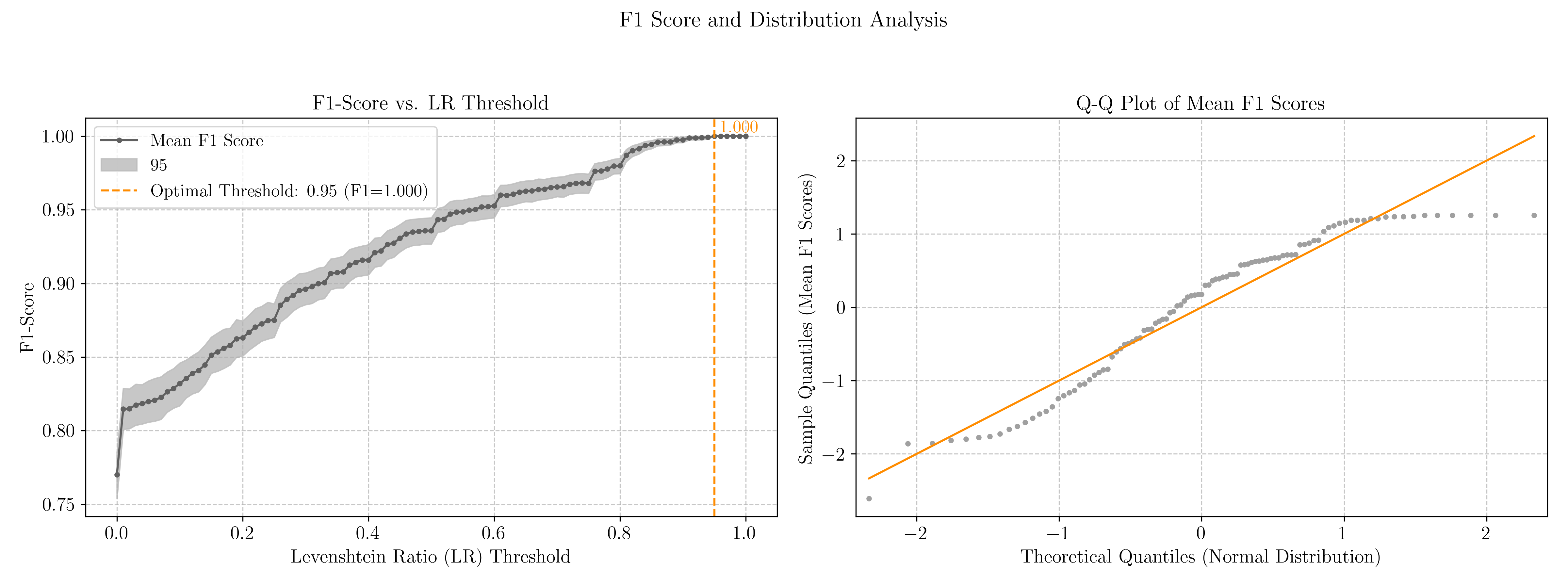}\label{fig:f1_vs_lr_threshold}
  \caption{F1-score versus Levenshtein Ratio (LR) threshold. The plot shows how the F1-score (comparing LR-based acceptance to EM ground truth) changes as the LR threshold is varied. The peak indicates the optimal LR threshold.}
\end{figure}

Figure~\ref{fig:f1_vs_lr_threshold} plots the F1-score against varying LR thresholds. The analysis indicated that an LR threshold of \textbf{0.95} maximized the F1-score (achieving an F1 of 1.000 in this specific analysis, suggesting very high agreement with EM at this tight threshold, while still allowing for very minor variations). Your initial analysis output confirms this optimal point. This empirically derived threshold provides a more robust alternative to an arbitrarily chosen one, such as the 0.75 threshold used for generating the LR Accuracy scores presented in the overview heatmap (Figure~\ref{fig:app_heatmap_lr_combo} in Appendix~\ref{app:lr_heatmap_figure}) and the detailed LR score table (Table~\ref{tab:app_detailed_lr_scores} in Appendix~\ref{app:detailed_scores}). The 0.75 threshold was used for those summary tables to align with common practice or initial exploratory settings before this specific F1-based calibration. This finding suggests that for the kind of short, factual answers typical of TQB\texorpdfstring{$^{++}$}{++}, a relatively high LR threshold (e.g., 0.95) is needed if the goal is to closely approximate EM while tolerating minimal differences. For practical purposes, an LR threshold between 0.85 and 0.95 might be a reasonable range depending on the desired trade-off between precision and recall.

It is also critical to reiterate that for multilingual LR calculations (especially with FR, JA, TR), Unicode normalization (e.g., to NFC or NFKC form) of both gold and predicted strings \textit{before} computing Levenshtein distance is essential to prevent spurious differences due to character encoding variations.
\FloatBarrier 

\subsection{Efficacy of Micro-Benchmarks}\label{sec:micro_efficacy}
A core proposition of TQB$^{++}$ is that even very small, synthetically generated datasets can serve as effective ``sense-checking'' or smoke tests. The experimental results support this. Performance trends observed on larger datasets (e.g., \texttt{core\_en} with 52 items or \texttt{pack\_en\_40} with 40 items) regarding model ranking and sensitivity to difficulty or language were largely mirrored on the smaller English packs like \texttt{pack\_en\_10} and \texttt{pack\_en\_20}.

For example, referring back to Figure~\ref{fig:heatmap_em_scores} and Appendix~\ref{app:lr_heatmap_figure}, the relative performance of different model families and sizes is generally discernible even with the 10-item \texttt{pack\_en\_10}. On the 10-item English pack (\texttt{pack\_en\_10}), larger models showcased strong performance. For instance, \texttt{gemma-3-12b} achieved an EM of 100.0, while \texttt{mistral-24b-instruct} scored EM 90.0. In comparison, smaller counterparts such as \texttt{llama-3.2-1b-instruct} scored EM 70.0, and \texttt{mistral-7b-instruct} obtained EM 60.0. More subtly, performance differences between variants of the same model family, such as \texttt{gemma-3-12b} versus \texttt{gemma-3-4b}, can also be detected. On \texttt{pack\_en\_10}, \texttt{gemma-3-12b} achieved an EM of 100.0, while \texttt{gemma-3-4b} scored 90.0. While the absolute scores might have higher variance with fewer items, the directional signal regarding severe performance regressions or basic capability checks, including clear differences between model sizes and more subtle ones between intra-family variants, remains.

The stability of EM scores across small English synthetic packs further illustrates this. For instance, \texttt{gemma-3-12b} scored EM 100.0 on \texttt{pack\_en\_10}, 90.0 on \texttt{pack\_en\_20}, and 93.3 on \texttt{pack\_en\_40}. Similarly, \texttt{mistral-24b-instruct} scored 90.0, 85.0, and 93.3 on these respective datasets. While not identical, these scores are comparable and show that a 10 or 20-item pack can provide a reasonable indication of performance that is not dramatically different from a 40-item pack for these types of general knowledge questions. This reinforces the idea that a small, well-constructed (even if synthetically generated) set can be surprisingly informative for quick checks, capable of surfacing both coarse and finer-grained performance signals.

This rapid feedback is invaluable in CI/CD pipelines or iterative prompt engineering, where waiting for extensive benchmark results is impractical. A 10 or 20-item TQB$^{++}$ pack can flag major issues in seconds.

\section{Related Work}\label{sec:related}
The evaluation of LLMs is a rapidly evolving field. TQB$^{++}$ situates itself within this landscape by addressing a specific need for lightweight, continuous testing.

\textbf{Large Benchmarks.} Comprehensive benchmarks like MMLU \citep{hendrycks2021measuringmassivemultitasklanguage}, BIG-Bench \citep{srivastava2023imitationgamequantifyingextrapolating}, and HELM \citep{liang2022helm} are essential for robustly assessing model capabilities across a wide array of tasks. However, their computational cost and time requirements make them unsuitable for high-frequency testing during development.

\textbf{Tiny and Efficient Benchmarks.} The need for more efficient evaluation has spurred research into smaller, targeted benchmarks. The tinyBenchmarks project \citep{polo2024tinybenchmarksevaluatingllmsfewer} demonstrates that carefully selected small subsets of major benchmarks can reliably approximate full benchmark performance for model ranking. TQB$^{++}$ shares this spirit of efficiency but targets an even smaller scale, focusing on CI smoke testing and regression detection rather than comparative model ranking. Other efforts like OpenAI Evals \citep{openai_evals} provide frameworks for running custom evaluations, and TQB can serve as a ready-made eval set for such systems.

\textbf{Synthetic QA Generation.} The generation of synthetic data for training and evaluation is a growing area. Surveys like \citet{long2024llmsdrivensyntheticdatageneration} cover the landscape of LLM-generated synthetic data. Some works focus on privacy-preserving synthetic data generation, such as differential privacy pipelines \citep{kurakin2024harnessinglargelanguagemodelsgenerate}. The TQB$^{++}$ generator, while also leveraging LLMs, emphasizes simplicity, rapid generation via minimal prompting, strict schema adherence for CI/CD compatibility, and provider-agnosticism through LiteLLM. This contrasts with other synthetic data efforts that might focus on generating highly complex reasoning chains, achieving state-of-the-art performance on specific benchmarks through synthetic training data, or involving more intricate generation and filtering pipelines. TQB$^{++}$'s approach is tailored for creating readily usable, schema-consistent micro-benchmarks for continuous evaluation and smoke testing.

\textbf{Tool‑Augmented QA and Agent Evaluation.} As LLMs are increasingly used as reasoning engines in autonomous agents that can use tools, benchmarks like ToolQA \citep{zhuang2023toolqadatasetllmquestion} are emerging to evaluate their ability to correctly use external tools. Furthermore, evolving comprehensive benchmarks such as AgentBench \citep{liu2023agentbenchevaluatingllmsagents}, which has incorporated increasingly complex tool-use sub-benchmarks, assess broader agent capabilities across diverse environments. TQB\texorpdfstring{$^{++}$}{++}, while not directly testing tool use or complex agency, can act as a preliminary gate: if an agent fails basic factual recall or reasoning on TQB, its more complex tool-use or autonomous capabilities might also be compromised.

\textbf{LLMOps and Continuous Evaluation.} The MLOps community has recognized the critical need for continuous monitoring of model quality. LLM observability tools and platforms \citep{signoz2024llmobservability, helicone2024prompteval} often integrate evaluation data. TQB was initially motivated by such use-cases, providing a concrete dataset for these frameworks.

\textbf{Dataset Inference and Provenance.} Concerns about data leakage and provenance in LLM training and evaluation are addressed by works like \citet{maini2021datasetinferenceownershipresolution}. While TQB is based on public domain facts, the TQB\texorpdfstring{$^{++}$}{++} generator's inclusion of SHA-256 hashes for synthetic items is a step towards better provenance for generated test sets. Furthermore, ensuring fairness and mitigating pre-existing biases in models themselves is a critical aspect of the LLMOps lifecycle, with dedicated frameworks proposed for their detection and mitigation \citep{Koc2025FrameworkFairnessML}.

\textbf{Metadata Standards for Datasets.} The Croissant initiative \citep{akhtar2024croissant} aims to standardize ML dataset metadata for better discovery and interoperability. TQB\texorpdfstring{$^{++}$}{++} adopts this standard to improve the usability of its artefacts.

\section{Conclusion}\label{sec:concl}
TQB\texorpdfstring{$^{++}$}{++} extends the original Tiny QA Benchmark by blending a deterministic 52‑item gold standard English dataset with an extensible synthetic data generation toolkit and machine‑readable Croissant metadata. The core TQB provides diverse basic QA coverage in a minimal footprint, ideal for rapid sanity checks. TQB\texorpdfstring{$^{++}$}{++} enhances this with multi‑lingual capabilities and custom test generation, aligning with modern LLMOps practices. It is designed to slot into CI/CD pipelines, prompt‑engineering workflows, and cross‑lingual deployment monitoring to catch regressions or logic bugs before more extensive and costly benchmarks are run. 

\paragraph{Limitations.} TQB\texorpdfstring{$^{++}$}{++} is intentionally designed as a lightweight smoke test focusing on factual QA. Consequently, its small scale and narrow task definition mean it cannot, by itself, detect broader issues such as complex instruction-following regressions, nuanced reasoning failures beyond factoid retrieval, or the generation of plausible-sounding but incorrect information (hallucinations) outside the specific QA context. Its utility is as a rapid first-pass filter, not a comprehensive measure of all desirable LLM capabilities.

All original code, the core dataset, Croissant files, and supplementary analysis materials are released under the Apache-2.0 license. Specific licensing details for all components, including synthetically generated datasets and the Python client library, are provided in Section~\ref{sec:ethics_license}. The resources are available on the Hugging Face Hub at \url{https://huggingface.co/datasets/vincentkoc/tiny_qa_benchmark_pp} and via a GitHub repository at \url{https://github.com/vincentkoc/tiny_qa_benchmark_pp}. The Python package for the generator can be installed from PyPI: \url{https://pypi.org/project/tinyqabenchmarkpp/}. We invite the community to utilize TQB\texorpdfstring{$^{++}$}{++}, fork the generator to add new domains and languages, and contribute to the development of robust, low-friction evaluation practices for LLMs.

\acks{We thank the anonymous reviewers for their insightful feedback. This work was supported in part by Comet ML, Inc. The author also acknowledges:
\begin{itemize}
    \item the use of OpenAI\'s `o3-mini' for generating some of the synthetic datasets used in the validation study.
    \item the use of OpenRouter for accessing the models used in the evaluation.
    \item the contributions of co-authors on the author's foundational publications in LLM evaluation, which have informed the direction of the current research.
\end{itemize}
}


\appendix
\section{Generation and Evaluation Protocol}\label{app:gen_eval_protocol}

This appendix details the specifics of the synthetic data generation process and the evaluation protocol used in the experiments described in Section~\ref{sec:experiments}.

\subsection{Synthetic Pack Generation Details}\label{app:generation_details}

The synthetic TQB\texorpdfstring{$^{++}$}{++} datasets were created using a Python script. (The script, \texttt{generate\_py}, is in the \texttt{suplimentary} directory included with this paper). The key aspects of the generation process are as follows:

\paragraph{Generator Script.} The script (approx. 300 lines of Python, and $<$200 when its not packaged for PyPi is leveraging \texttt{litellm}) utilizes the \texttt{litellm} library to interact with various LLM providers. It takes parameters such as the number of questions, target language(s), categories, difficulty, and the specific LLM provider and model to use for generation.

\paragraph{Prompting Strategy.} A structured prompting approach was employed:
\begin{itemize}
    \item \textbf{System Prompt:} ``You are a dataset generator that outputs JSON with keys: text, label, context, tags{category, difficulty[easy, medium, hard]}. The context MUST be a one-sentence fact that contains the label verbatim." An optional domain context can be appended if provided by the user.
    \item \textbf{Few-shot Examples:} Two examples are provided to the model within the user message to guide the output format and style:
    Example 1:
    \begin{lstlisting}[basicstyle=\ttfamily\footnotesize, breaklines=true, columns=fullflexible,numbers=none]
{
    "text": "What is 2 + 2?",
    "label": "4",
    "context": "2 + 2 equals 4.",
    "tags": {
        "category": "math",
        "difficulty": "easy"
    }
}
    \end{lstlisting}
    Example 2:
    \begin{lstlisting}[basicstyle=\ttfamily\footnotesize, breaklines=true, columns=fullflexible,numbers=none]
{
    "text": "Who wrote '1984'?",
    "label": "George Orwell",
    "context": "The novel 1984 was written by George Orwell.",
    "tags": {
        "category": "literature",
        "difficulty": "easy"
    }
}
    \end{lstlisting}
    \item \textbf{User Prompt:} A template like ``Generate {n} {diff} questions in {lang} about {category}. Return a JSON list ONLY." is used, filled with the user-specified parameters.
\end{itemize}

\paragraph{Generation Model and Hyperparameters.} 
For generating all synthetic datasets used in this paper's experiments, the following setup was used:
\begin{itemize}
    \item \textbf{Model:} OpenAI's \texttt{gpt-3.5-turbo-0125} (referred to as \texttt{o3-mini} in Section~\ref{sec:experiments}), accessed via an OpenAI-compatible API endpoint.
    \item \textbf{Temperature:} 1.0 as its set for OpenAI reasoning models, otherwise would use 0.0 (as mentioned in Section~\ref{sec:experiments}, to increase reprodceability in generation).
    \item \textbf{Max Tokens:} 4096 (default for the generation script, \texttt{generator.py}).
    \item \textbf{Seed:} While the script supports a seed argument for providers that use it, specific seed values were generally used during batch generation runs to aim for reproducibility where possible, though a fixed seed was not strictly enforced for every single generation instance if not supported by a particular endpoint or if exploratory generation was performed. For the reported benchmark packs, best efforts were made to use consistent generation parameters.
    \item \textbf{Response Format:} JSON object mode was requested from the API where available (e.g., \texttt{\{"type": "json\_object"\}}).
\end{itemize}

\paragraph{Post-filtering and Validation.} 
The primary validation performed by the \texttt{generate\_py} script is schema adherence. It checks if the LLM output is valid JSON and if the expected keys (\texttt{text}, \texttt{label}, \texttt{context}, \texttt{tags} with sub-keys \texttt{category} and \texttt{difficulty}) are present in each generated item. If the output is not valid JSON or a list of items, it may attempt to parse content from provider-specific refusal fields or retry (though retry logic was not heavily relied upon for the main benchmark generation). No further automated post-filtering, such as self-consistency checks or external factuality verification against a knowledge base, is implemented directly within this basic generation script. The quality relies on the LLM's ability to follow instructions and the schema validation. A SHA-256 hash is computed for each generated item for provenance.

\subsection{Evaluation Protocol}\label{app:evaluation_protocol}

\paragraph{Models Evaluated.} The models used in the evaluation (Section~\ref{sec:results}) were accessed via LiteLLM, typically through an OpenRouter endpoint or directly if applicable. Table~\ref{tab:app_models_evaluated} lists the models, their providers/full IDs used in \texttt{eval\_batch\_py}, and relevant notes.

\begin{table*}[htbp!]
  \centering
  \caption{Models Used in Evaluation.}\label{tab:app_models_evaluated}
  \footnotesize
  \begin{tabular}{@{}ll@{}}
    \toprule
    Model Alias (in Paper) & Provider Notes \\
    \midrule
    \texttt{gemma-3-12b} & OpenRouter \\
    \texttt{gemma-3-4b}  & OpenRouter \\
    \texttt{ministral-8b} & OpenRouter \\
    \texttt{ministral-3b} & OpenRouter \\
    \texttt{mistral-24b-instruct} & OpenRouter \\
    \texttt{mistral-7b-instruct} & OpenRouter \\
    \texttt{llama-3.2-3b-instruct} & OpenRouter \\
    \texttt{llama-3.2-1b-instruct} & OpenRouter \\
    \bottomrule
  \end{tabular}
  \caption*{\footnotesize Note: Specific API access costs were not tracked for this study, but choices generally favored accessible models. Latency can vary based on provider and real-time load.}
\end{table*}

OpenAI models were not used for the evaluation runs presented in Section~\ref{sec:results}. This decision was due to the general lack of publicly disclosed model parameters and comprehensive model cards for OpenAI models, which hinders comparative analysis across model families with varying weights. Additionally, this choice helps to mitigate potential self-reinforcement bias, as an OpenAI model was used for the synthetic data generation.

\paragraph{Rationale for Generation Model Choice (\texttt{o3-mini}).} 
The synthetic datasets were primarily generated using OpenAI's \texttt{gpt-3.5-turbo-0125} model (aliased as \texttt{o3-mini}). This choice was based on a pragmatic balance of factors: 
\begin{itemize}
    \item \textbf{Accessibility and Cost:} OpenAI models are widely accessible via API. For instance, the \texttt{gpt-3.5-turbo} series offers a good balance of capability versus cost for generating moderately sized datasets.
    \item \textbf{Reasoning Capability for Quality Data:} To ensure the generated synthetic data was of reasonable quality, a model with strong reasoning and instruction-following capabilities was preferred. While more advanced models like \texttt{GPT-4} or Claude 3 Opus might offer higher quality, \texttt{gpt-3.5-turbo} variants are often considered highly performant for structured data generation tasks (as also indicated by general leaderboards like those on the Artificial Analysis website \citep{ArtificialAnalysisWeb}). Using a capable model was deemed important for the integrity of the benchmark, even if it was not the absolute top-tier model, to ensure the questions and answers were coherent and factually plausible for a smoke test.
    \item \textbf{Speed of Generation:} For rapidly creating multiple dataset variants across languages and categories, a model with reasonable generation speed was also a factor.
\end{itemize}
Alternative high-quality reasoning models like Google's Gemini 1.5 Pro or Anthropic's Claude series could also be suitable candidates for generation, contingent on API access, cost, and specific feature requirements (e.g., JSON mode support).

\paragraph{Evaluation Script and Hyperparameters.} All evaluations were conducted using the \texttt{eval\_py} script. For these runs, the temperature was set to 0.0 (for OpenAI reasoning models we had to use 1.0) to ensure deterministic outputs from the models under test, and a consistent seed (defaulting to 42 in the script if not overridden) was used for reproducibility where supported by the LLM provider and model.

\vspace{0.5\baselineskip}
\section{Additional Experimental Results}\label{app:additional_results}

\subsection{Levenshtein Ratio (LR) Score Heatmap}\label{app:lr_heatmap_figure}
\begin{figure}[htb] 
  \centering
  \begin{minipage}[t]{0.48\linewidth}
    \centering
    \includegraphics[width=\linewidth]{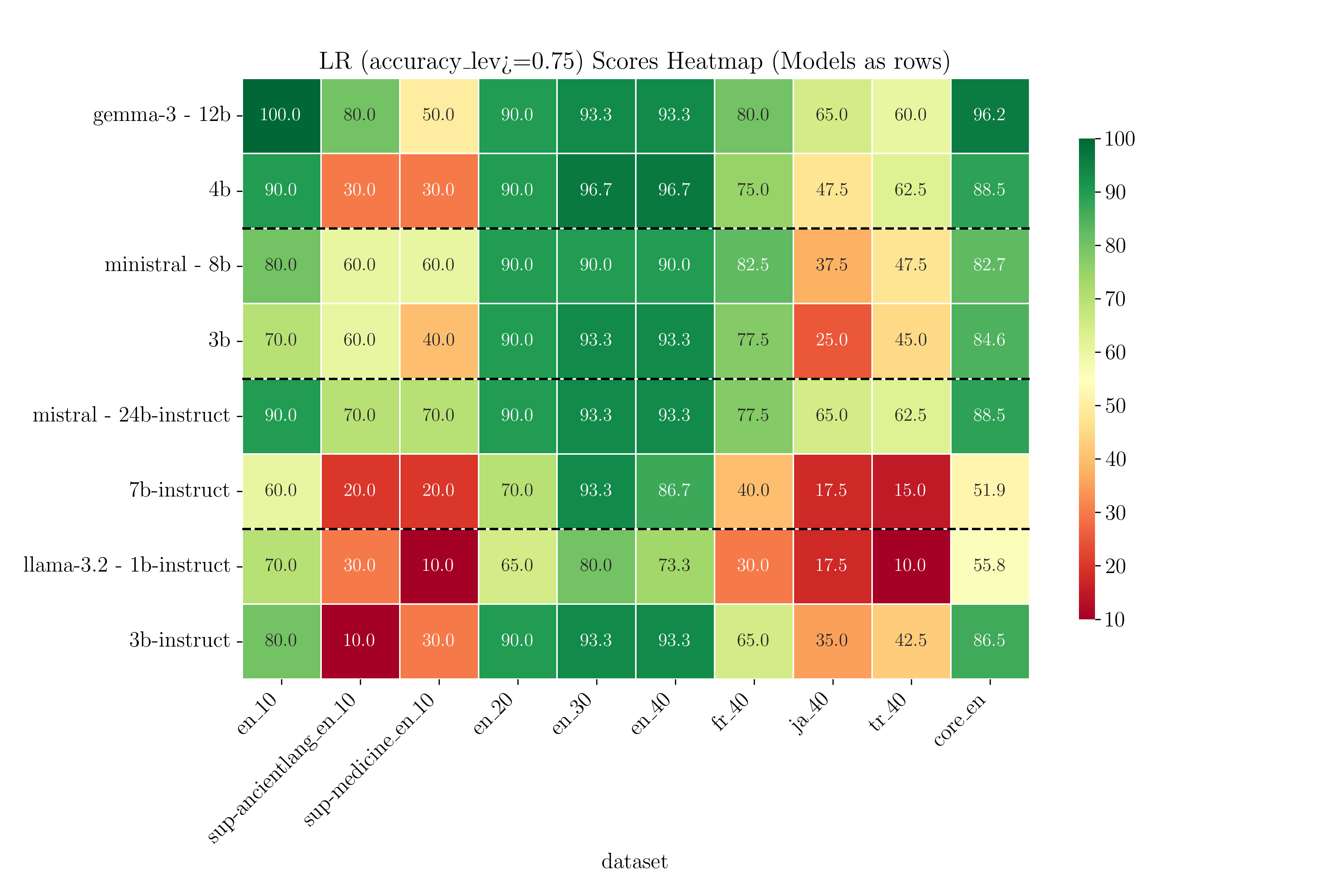}
    \caption{LR Scores by Model/Dataset.}
    \label{fig:app_heatmap_lr_combo}
  \end{minipage}\hfill
  \begin{minipage}[t]{0.48\linewidth}
    \centering
    \includegraphics[width=\linewidth]{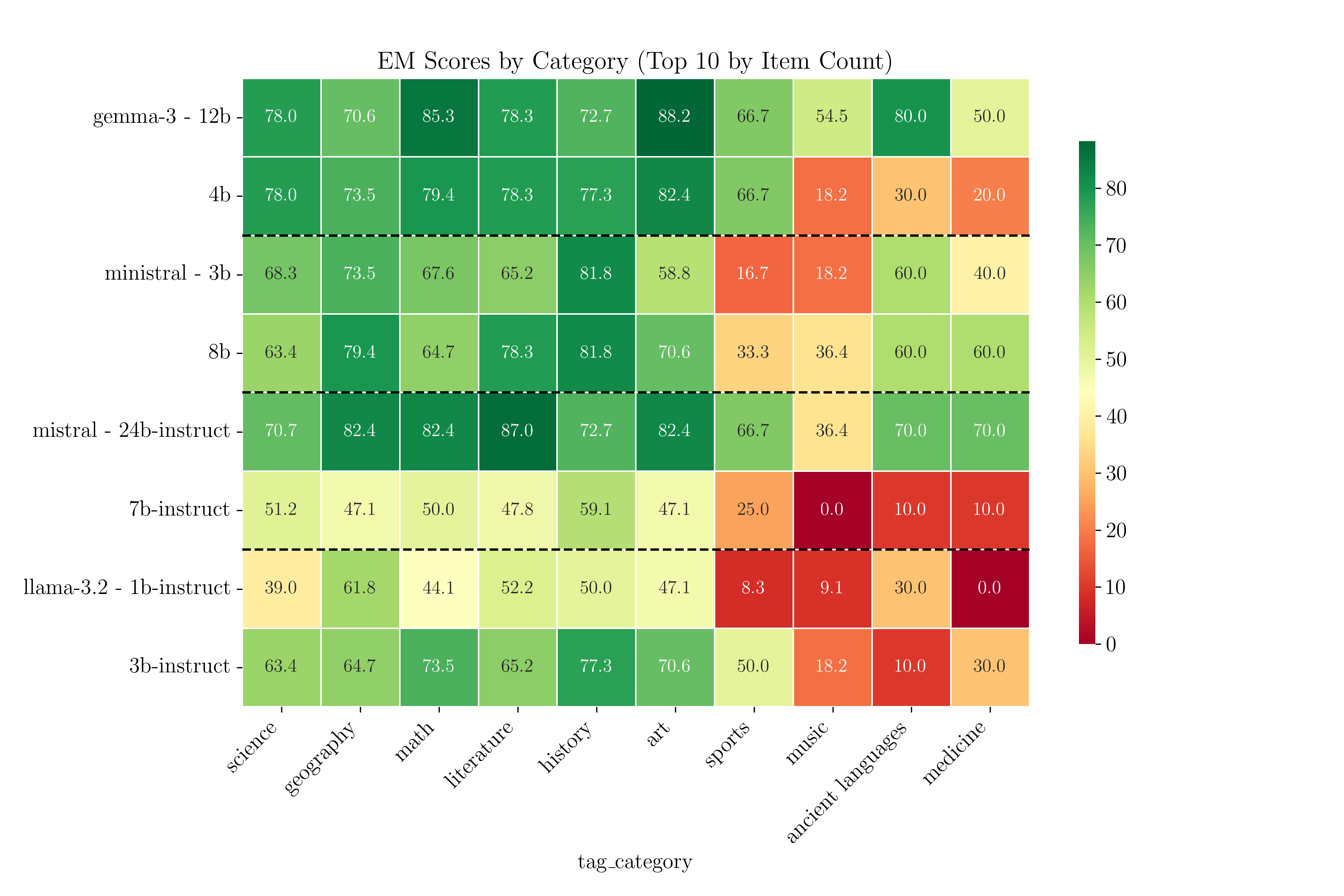}
    \caption{EM Scores by Model/Category.}
    \label{fig:app_heatmap_category_combo}
  \end{minipage}
  \caption*{\footnotesize Figure~\ref{fig:app_heatmap_lr_combo} (left) shows Levenshtein Ratio (LR) scores (threshold $\ge0.75$). Figure~\ref{fig:app_heatmap_category_combo} (right) shows mean Exact Match (EM) scores by category. Darker shades indicate higher accuracy.}
\end{figure}

Figure~\ref{fig:app_heatmap_lr_combo} shows the Levenshtein Ratio (LR) scores, using an LR threshold $\ge0.75$ as discussed in Section~\ref{sec:results}. 

\subsection{Category Score Heatmap}\label{app:category_heatmap_figure}
The full heatmap of mean EM scores by model and category is shown in Figure~\ref{fig:app_heatmap_category_combo}.

\section{Detailed Score Matrix with Intra-Family Deltas}\label{app:detailed_scores}
This appendix presents comprehensive score matrices for all evaluated models across the primary TQB and TQB\texorpdfstring{$^{++}$}{++} datasets (excluding the supplementary challenge datasets: \texttt{sup-ancientlang\_en\_10} and \texttt{sup-medicine\_en\_10}, which are, however, included in overview Figure~\ref{fig:heatmap_em_scores}). For model pairs within the same family, a $\Delta$ row is included to highlight the performance difference between the larger and smaller variant on each dataset. 

Table~\ref{tab:app_detailed_em_scores} presents the detailed Exact Match (EM) scores, where larger model variants generally outperform smaller ones within the same family, though with some variance by language and dataset size.
{\footnotesize 
\begin{table*}[h]
  \centering
  \caption{Detailed Exact Match (EM) Scores (0-100 Scale) by Model and Dataset (Supplementary Datasets Excluded), with Intra-Family $\Delta$.}\label{tab:app_detailed_em_scores}
  \footnotesize
  \begin{tabular}{@{}llcccccccc@{}}
    \toprule
    Model Family & Model Variant & \texttt{en\_10} & \texttt{en\_20} & \texttt{en\_30} & \texttt{en\_40} & \texttt{fr\_40} & \texttt{ja\_40} & \texttt{tr\_40} & \texttt{core\_en} \\
    \midrule
    Gemma-3 & \texttt{gemma-3-12b} & 100.0 & 90.0 & 93.3 & 93.3 & 80.0 & 50.0 & 50.0 & 90.4 \\
            & \texttt{gemma-3-4b} & 90.0 & 90.0 & 96.7 & 96.7 & 75.0 & 37.5 & 50.0 & 86.5 \\
            & \textit{$\Delta$ (12b-4b)} & \textit{10.0} & \textit{0.0} & \textit{-3.4} & \textit{-3.4} & \textit{5.0} & \textit{12.5} & \textit{0.0} & \textit{3.9} \\
    \midrule
    Ministral & \texttt{ministral-8b} & 80.0 & 90.0 & 90.0 & 90.0 & 72.5 & 35.0 & 37.5 & 80.8 \\
              & \texttt{ministral-3b} & 70.0 & 90.0 & 93.3 & 93.3 & 65.0 & 20.0 & 40.0 & 76.9 \\
              & \textit{$\Delta$ (8b-3b)} & \textit{10.0} & \textit{0.0} & \textit{-3.3} & \textit{-3.3} & \textit{7.5} & \textit{15.0} & \textit{-2.5} & \textit{3.9} \\
    \midrule
    Mistral & \texttt{mistral-24b-instruct} & 90.0 & 85.0 & 93.3 & 93.3 & 72.5 & 47.5 & 55.0 & 84.6 \\
            & \texttt{mistral-7b-instruct} & 60.0 & 70.0 & 90.0 & 83.3 & 32.5 & 7.5 & 12.5 & 50.0 \\
            & \textit{$\Delta$ (24b-7b)} & \textit{30.0} & \textit{15.0} & \textit{3.3} & \textit{10.0} & \textit{40.0} & \textit{40.0} & \textit{42.5} & \textit{34.6} \\
    \midrule
    Llama-3.2 & \texttt{llama-3.2-3b-instruct} & 80.0 & 90.0 & 93.3 & 93.3 & 57.5 & 22.5 & 37.5 & 84.6 \\
              & \texttt{llama-3.2-1b-instruct} & 70.0 & 65.0 & 80.0 & 73.3 & 25.0 & 12.5 & 7.5 & 53.8 \\
              & \textit{$\Delta$ (3b-1b)} & \textit{10.0} & \textit{25.0} & \textit{13.3} & \textit{20.0} & \textit{32.5} & \textit{10.0} & \textit{30.0} & \textit{30.8} \\
    \bottomrule
  \end{tabular}
\end{table*}
}

Levenshtein Ratio (LR) scores (Table~\ref{tab:app_detailed_lr_scores}), using a $\ge0.75$ threshold, show a similar trend, often with slightly higher scores than EM due to partial credit for near matches.
{\footnotesize 
\begin{table*}[h]
  \centering
  \caption{Detailed Levenshtein Ratio (LR) Scores (0-100 Scale, using LR threshold $\ge0.75$) by Model and Dataset (Supplementary Datasets Excluded), with Intra-Family $\Delta$.}\label{tab:app_detailed_lr_scores}
  \footnotesize
  \begin{tabular}{@{}llcccccccc@{}}
    \toprule
    Model Family & Model Variant & \texttt{en\_10} & \texttt{en\_20} & \texttt{en\_30} & \texttt{en\_40} & \texttt{fr\_40} & \texttt{ja\_40} & \texttt{tr\_40} & \texttt{core\_en} \\
    \midrule
    Gemma-3 & \texttt{gemma-3-12b} & 100.0 & 90.0 & 93.3 & 93.3 & 80.0 & 65.0 & 60.0 & 96.2 \\
            & \texttt{gemma-3-4b} & 90.0 & 90.0 & 96.7 & 96.7 & 75.0 & 47.5 & 62.5 & 88.5 \\
            & \textit{$\Delta$ (12b-4b)} & \textit{10.0} & \textit{0.0} & \textit{-3.4} & \textit{-3.4} & \textit{5.0} & \textit{17.5} & \textit{-2.5} & \textit{7.7} \\
    \midrule
    Ministral & \texttt{ministral-8b} & 80.0 & 90.0 & 90.0 & 90.0 & 82.5 & 37.5 & 47.5 & 82.7 \\
              & \texttt{ministral-3b} & 70.0 & 90.0 & 93.3 & 93.3 & 77.5 & 25.0 & 45.0 & 84.6 \\
              & \textit{$\Delta$ (8b-3b)} & \textit{10.0} & \textit{0.0} & \textit{-3.3} & \textit{-3.3} & \textit{5.0} & \textit{12.5} & \textit{2.5} & \textit{-1.9} \\
    \midrule
    Mistral & \texttt{mistral-24b-instruct} & 90.0 & 90.0 & 93.3 & 93.3 & 77.5 & 65.0 & 62.5 & 88.5 \\
            & \texttt{mistral-7b-instruct} & 60.0 & 70.0 & 93.3 & 86.7 & 40.0 & 17.5 & 15.0 & 51.9 \\
            & \textit{$\Delta$ (24b-7b)} & \textit{30.0} & \textit{20.0} & \textit{0.0} & \textit{6.6} & \textit{37.5} & \textit{47.5} & \textit{47.5} & \textit{36.6} \\
    \midrule
    Llama-3.2 & \texttt{llama-3.2-3b-instruct} & 80.0 & 90.0 & 93.3 & 93.3 & 65.0 & 35.0 & 42.5 & 86.5 \\
              & \texttt{llama-3.2-1b-instruct} & 70.0 & 65.0 & 80.0 & 73.3 & 30.0 & 17.5 & 10.0 & 55.8 \\
              & \textit{$\Delta$ (3b-1b)} & \textit{10.0} & \textit{25.0} & \textit{13.3} & \textit{20.0} & \textit{35.0} & \textit{17.5} & \textit{32.5} & \textit{30.7} \\
    \bottomrule
  \end{tabular}
\end{table*}
}

The difference between LR and EM scores (Table~\ref{tab:app_detailed_lrem_diff}) highlights how much each model benefits from the more lenient LR metric, often indicating generation of semantically close but not identical answers, particularly in non-English languages.
{\footnotesize 
\begin{table*}[h]
  \centering
  \caption{Difference Between Detailed LR and EM Scores (LR - EM) by Model and Dataset (Supplementary Datasets Excluded).}\label{tab:app_detailed_lrem_diff}
  \footnotesize
  \begin{tabular}{@{}llcccccccc@{}}
    \toprule
    Model Family & Model Variant & \texttt{en\_10} & \texttt{en\_20} & \texttt{en\_30} & \texttt{en\_40} & \texttt{fr\_40} & \texttt{ja\_40} & \texttt{tr\_40} & \texttt{core\_en} \\
    \midrule
    Gemma-3 & \texttt{gemma-3-12b} & 0.0 & 0.0 & 0.0 & 0.0 & 0.0 & 15.0 & 10.0 & 5.8 \\
            & \texttt{gemma-3-4b} & 0.0 & 0.0 & 0.0 & 0.0 & 0.0 & 10.0 & 12.5 & 2.0 \\
    \midrule
    Ministral & \texttt{ministral-8b} & 0.0 & 0.0 & 0.0 & 0.0 & 10.0 & 2.5 & 10.0 & 1.9 \\
              & \texttt{ministral-3b} & 0.0 & 0.0 & 0.0 & 0.0 & 12.5 & 5.0 & 5.0 & 7.7 \\
    \midrule
    Mistral & \texttt{mistral-24b-instruct} & 0.0 & 5.0 & 0.0 & 0.0 & 5.0 & 17.5 & 7.5 & 3.9 \\
            & \texttt{mistral-7b-instruct} & 0.0 & 0.0 & 3.3 & 3.4 & 7.5 & 10.0 & 2.5 & 1.9 \\
    \midrule
    Llama-3.2 & \texttt{llama-3.2-3b-instruct} & 0.0 & 0.0 & 0.0 & 0.0 & 7.5 & 12.5 & 5.0 & 1.9 \\
              & \texttt{llama-3.2-1b-instruct} & 0.0 & 0.0 & 0.0 & 0.0 & 5.0 & 5.0 & 2.5 & 2.0 \\
    \bottomrule
  \end{tabular}
\end{table*}
}
\FloatBarrier

\section{Sample Dataset Items}\label{app:sample_items}
\normalsize 
This section provides a few illustrative examples from the core TQB dataset and some of the generated TQB\texorpdfstring{$^{++}$}{++} packs to give a qualitative sense of the data.

\subsection*{Sample Dataset Items}
\lstset{basicstyle=\ttfamily\footnotesize,breaklines=true,columns=fullflexible} 

\subsubsection*{\texttt{core\_en.json} (Human-Curated English Core)}
\begin{lstlisting}
{"text":"What is the capital of France?","label":"Paris","metadata":{"context":"France is a country in Europe. Its capital is Paris."},"tags":{"category":"geography","difficulty":"easy"}}
\end{lstlisting}

\subsubsection*{\texttt{pack\_en\_10.json} (Synthetically Generated English, n=10)}
\begin{lstlisting}
{"text":"In which year did the Titanic sink?","label":"1912","context":"The Titanic sank in 1912.","tags":{"category":"history","difficulty":"medium"},"id":"ecfc9a30","lang":"en","sha256":"5599c977...40bcf6a1"}
\end{lstlisting}

\subsubsection*{\texttt{pack\_fr\_40.json} (Synthetically Generated French, n=40)}
\begin{lstlisting}
{"text":"Combien font 5 + 7 ?","label":"12","context":"5 + 7 = 12.","tags":{"category":"math","difficulty":"easy"},"id":"292402c2","lang":"fr","sha256":"762e734d...b8b6085"}
\end{lstlisting}

\lstset{basicstyle=\ttfamily\small,breaklines=true} 

\vskip 0.2in 
\bibliography{tinyqa_pp_full} 

\begin{thebibliography}{23}
\providecommand{\natexlab}[1]{#1}
\providecommand{\url}[1]{\texttt{#1}}
\expandafter\ifx\csname urlstyle\endcsname\relax
  \providecommand{\doi}[1]{doi: #1}\else
  \providecommand{\doi}{doi: \begingroup \urlstyle{rm}\Url}\fi

\bibitem[Akhtar et~al.(2024)Akhtar, de~Castro, Egele, endangering Trazona, Massonnet, Moir, Kanter, Vanschoren, Pagels, Hudecek, Zaldivar, Kim, and Passat]{akhtar2024croissant}
Omar Akhtar, Ruanne de~Castro, Romain Egele, Osma endangering Trazona, Stephane Massonnet, Sarah Moir, David Kanter, Joaquin Vanschoren, Max Pagels, Vojtech Hudecek, Andrew Zaldivar, Been Kim, and Nicolas Passat.
\newblock Croissant: A metadata format for machine learning datasets.
\newblock \emph{arXiv preprint arXiv:2401.12982}, 2024.

\bibitem[{Artificial Analysis}(2024)]{ArtificialAnalysisWeb}
{Artificial Analysis}.
\newblock Independent analysis of ai models and api providers.
\newblock \url{https://artificialanalysis.ai/}, 2024.
\newblock Accessed: 2025-05-15.

\bibitem[Comet~ML(2025)]{cometopikoptimizerdocs}
Inc. Comet~ML.
\newblock Cometopik optimizer documentation.
\newblock \url{https://www.comet.com/docs/opik/agent_optimization/opik_optimizer}, 2025.
\newblock Accessed: 2025-05-15.

\bibitem[Hendrycks et~al.(2021)Hendrycks, Burns, Basart, Zou, Mazeika, Song, and Steinhardt]{hendrycks2021measuringmassivemultitasklanguage}
Dan Hendrycks, Collin Burns, Steven Basart, Andy Zou, Mantas Mazeika, Dawn Song, and Jacob Steinhardt.
\newblock Measuring massive multitask language understanding.
\newblock \emph{arXiv preprint arXiv:2009.03300}, 2021.
\newblock URL \url{https://arxiv.org/abs/2009.03300}.

\bibitem[Hochlehnert et~al.(2025)Hochlehnert, Bhatnagar, Udandarao, Albanie, Prabhu, and Bethge]{hochlehnert2025soberlookprogresslanguage}
Andreas Hochlehnert, Hardik Bhatnagar, Vishaal Udandarao, Samuel Albanie, Ameya Prabhu, and Matthias Bethge.
\newblock A sober look at progress in language model reasoning: Pitfalls and paths to reproducibility, 2025.
\newblock URL \url{https://arxiv.org/abs/2504.07086}.

\bibitem[J(2024)]{signoz2024llmobservability}
Jaikanth J.
\newblock Understanding llm observability - key insights, best practices, \& tools.
\newblock \url{https://signoz.io/blog/llm-observability/}, September 2024.
\newblock Accessed: 2025-05-15.

\bibitem[Koc(2025{\natexlab{a}})]{Koc2025FrameworkFairnessML}
V.~Koc.
\newblock Framework for fairness in machine learning using detecting and mitigating bias in ai algorithms.
\newblock In \emph{2025 3rd IEEE International Conference on Business Analytics for Technology and Security (ICBATS-2025)}, Dubai, United Arab Emirates, May 2025{\natexlab{a}}.

\bibitem[Koc et~al.(2025)Koc, Alang, Janjua, and Peta]{Koc2025LeveragingMultipleLLMEvaluators}
V.~Koc, K.~Alang, J.~I. Janjua, and S.~B. Peta.
\newblock Leveraging multiple llm evaluators for scalable and fair language model assessments.
\newblock In \emph{2025 IEEE International Conference on Metaverse and Current Trends in Computing (ICMCTC)}, Subang Jaya, Malaysia, April 2025.

\bibitem[Koc(2025{\natexlab{b}})]{Koc202XNonLatinLLMEvalComet}
Vincent Koc.
\newblock Complexities for non-latin languages \& llm evaluations, 03 2025{\natexlab{b}}.
\newblock URL \url{https://www.comet.com/site/blog/complexities-for-non-latin-languages-llm-evaluations/}.
\newblock Accessed on the urldate.

\bibitem[Koc(2025{\natexlab{c}})]{koc2025generativeailargelanguage}
Vincent Koc.
\newblock Generative ai and large language models in language preservation: Opportunities and challenges, 01 2025{\natexlab{c}}.
\newblock URL \url{https://arxiv.org/abs/2501.11496}.

\bibitem[Koc(2025{\natexlab{d}})]{koctinyqabenchmark_original}
Vincent Koc.
\newblock tiny\_qa\_benchmark (revision ff9143f), 04 2025{\natexlab{d}}.
\newblock URL \url{https://huggingface.co/datasets/vincentkoc/tiny_qa_benchmark}.

\bibitem[Koc(2025{\natexlab{e}})]{koctinyqabenchmarkpp}
Vincent Koc.
\newblock Tiny qa benchmark++ (tqb++) datasets and toolkit.
\newblock \url{https://huggingface.co/datasets/vincentkoc/tiny_qa_benchmark_pp}, 2025{\natexlab{e}}.
\newblock See also: \url{https://github.com/vincentkoc/tiny_qa_benchmark_pp}.

\bibitem[Kurakin et~al.(2024)Kurakin, Ponomareva, Syed, MacDermed, and Terzis]{kurakin2024harnessinglargelanguagemodelsgenerate}
Alexey Kurakin, Natalia Ponomareva, Umar Syed, Liam MacDermed, and Andreas Terzis.
\newblock Harnessing large-language models to generate private synthetic text, 2024.
\newblock URL \url{https://arxiv.org/abs/2306.01684}.

\bibitem[Lam(2024)]{helicone2024prompteval}
Lina Lam.
\newblock The ultimate guide to prompt evaluation frameworks.
\newblock \url{https://www.helicone.ai/blog/prompt-evaluation-frameworks}, 2024.
\newblock Accessed: 2025-05-15.

\bibitem[Levenshtein(1966)]{levenshtein1966binary}
Vladimir~I. Levenshtein.
\newblock Binary codes capable of correcting deletions, insertions, and reversals.
\newblock \emph{Soviet Physics Doklady}, 10\penalty0 (8):\penalty0 707--710, 1966.

\bibitem[Liang et~al.(2022)Liang, Bommasani, Lee, Tsipras, Soylu, Yasunaga, Zhang, Narayanan, Wu, Kumar, et~al.]{liang2022helm}
Percy Liang, Rishi Bommasani, Tony Lee, Dimitris Tsipras, Dilara Soylu, Michihiro Yasunaga, Yian Zhang, Deepak Narayanan, Yuhuai Wu, Ananya Kumar, et~al.
\newblock Holistic evaluation of language models.
\newblock \emph{arXiv preprint arXiv:2211.09110}, 2022.

\bibitem[Liu et~al.(2023)Liu, Yu, Zhang, Xu, Lei, Lai, Gu, Ding, Men, Yang, Zhang, Deng, Zeng, Du, Zhang, Shen, Zhang, Su, Sun, Huang, Dong, and Tang]{liu2023agentbenchevaluatingllmsagents}
Xiao Liu, Hao Yu, Hanchen Zhang, Yifan Xu, Xuanyu Lei, Hanyu Lai, Yu~Gu, Hangliang Ding, Kaiwen Men, Kejuan Yang, Shudan Zhang, Xiang Deng, Aohan Zeng, Zhengxiao Du, Chenhui Zhang, Sheng Shen, Tianjun Zhang, Yu~Su, Huan Sun, Minlie Huang, Yuxiao Dong, and Jie Tang.
\newblock Agentbench: Evaluating llms as agents, 2023.
\newblock URL \url{https://arxiv.org/abs/2308.03688}.

\bibitem[Long et~al.(2024)Long, Wang, Xiao, Zhao, Ding, Chen, and Wang]{long2024llmsdrivensyntheticdatageneration}
Lin Long, Rui Wang, Ruixuan Xiao, Junbo Zhao, Xiao Ding, Gang Chen, and Haobo Wang.
\newblock On llms-driven synthetic data generation, curation, and evaluation: A survey, 2024.
\newblock URL \url{https://arxiv.org/abs/2406.15126}.

\bibitem[Maini et~al.(2021)Maini, Yaghini, and Papernot]{maini2021datasetinferenceownershipresolution}
Pratyush Maini, Mohammad Yaghini, and Nicolas Papernot.
\newblock Dataset inference: Ownership resolution in machine learning, 2021.
\newblock URL \url{https://arxiv.org/abs/2104.10706}.

\bibitem[{OpenAI}(2024)]{openai_evals}
{OpenAI}.
\newblock Openai evals framework.
\newblock \url{https://github.com/openai/evals}, 2024.
\newblock Accessed: 2025-05-15.

\bibitem[Polo et~al.(2024)Polo, Weber, Choshen, Sun, Xu, and Yurochkin]{polo2024tinybenchmarksevaluatingllmsfewer}
Felipe~Maia Polo, Lucas Weber, Leshem Choshen, Yuekai Sun, Gongjun Xu, and Mikhail Yurochkin.
\newblock tinybenchmarks: evaluating llms with fewer examples, 2024.
\newblock URL \url{https://arxiv.org/abs/2402.14992}.

\bibitem[Srivastava et~al.(2023)]{srivastava2023imitationgamequantifyingextrapolating}
Aarohi Srivastava et~al.
\newblock Beyond the imitation game: Quantifying and extrapolating the capabilities of language models, 2023.
\newblock URL \url{https://arxiv.org/abs/2206.04615}.

\bibitem[Zhuang et~al.(2023)Zhuang, Yu, Wang, Sun, and Zhang]{zhuang2023toolqadatasetllmquestion}
Yuchen Zhuang, Yue Yu, Kuan Wang, Haotian Sun, and Chao Zhang.
\newblock Toolqa: A dataset for llm question answering with external tools, 2023.
\newblock URL \url{https://arxiv.org/abs/2306.13304}.

\end{thebibliography}

\end{document}